\documentclass[10pt,twocolumn,letterpaper]{article}

\usepackage{cvpr}
\usepackage{times}
\usepackage{epsfig}
\usepackage{graphicx}
\usepackage{amsmath}
\usepackage{amssymb}
\usepackage[numbers,sort&compress]{natbib} 
\usepackage{multirow}

\usepackage{xcolor,tabulary,xfrac,enumitem}
\usepackage[pagebackref=true,breaklinks=true,letterpaper=true,colorlinks,citecolor=citecolor,bookmarks=false]{hyperref}
\definecolor{citecolor}{RGB}{34,139,34}
\usepackage{booktabs}

\cvprfinalcopy 

\def\cvprPaperID{****} 

\ifcvprfinal\fi
\begin{document}

\title{LID 2020: The Learning from Imperfect Data Challenge Results}

\author{Yunchao Wei, Shuai Zheng, Ming-Ming Cheng, Hang Zhao, Liwei Wang, Errui Ding, Yi Yang, \\
Antonio Torralba, Ting Liu, Guolei Sun, Wenguan Wang, Luc Van Gool, Wonho Bae, Junhyug Noh,\\ Jinhwan Seo, Gunhee Kim, Hao Zhao, Ming Lu, Anbang Yao, Yiwen Guo, Yurong Chen, Li Zhang,\\ Chuangchuang Tan, Tao Ruan, Guanghua Gu, Shikui Wei, Yao Zhao, Mariia Dobko, Ostap Viniavskyi,\\ Oles Dobosevych, Zhendong Wang, Zhenyuan Chen, Chen Gong, Huanqing Yan, Jun He\\

}
\maketitle

\begin{abstract}
Learning from imperfect data becomes an issue in many industrial applications after the research community has made profound progress in supervised learning from perfectly annotated datasets. The purpose of the Learning from Imperfect Data (LID) workshop is to inspire and facilitate the research in developing novel approaches that would harness the imperfect data and improve the data-efficiency during training. A massive amount of user-generated data nowadays available on multiple internet services. How to leverage those and improve the machine learning models is a high impact problem. We organize the challenges in conjunction with the workshop. The goal of these challenges is to find the state-of-the-art approaches in the weakly supervised learning setting for object detection, semantic segmentation, and scene parsing. There are three tracks in the challenge, \ie, weakly supervised semantic segmentation (Track 1), weakly supervised scene parsing (Track 2), and weakly supervised object localization (Track 3). In Track 1, based on ILSVRC DET~\cite{2009-imagenet}, we provide pixel-level annotations of 15K images from 200 categories for evaluation. In Track 2, we provide point-based annotations for the training set of ADE20K~\cite{zhou2017scene}. In Track 3, based on ILSVRC CLS-LOC~\cite{2009-imagenet}, we provide pixel-level annotations of 44,271 images for evaluation~\cite{zhang2020rethinking}. Besides, we further introduce a new evaluation metric proposed by~\cite{zhang2020rethinking}, \ie, IoU curve, to measure the quality of the generated object localization maps. This technical report summarizes the highlights from the challenge. The challenge submission server and the leaderboard will continue to open for the researchers who are interested in it. More details regarding the challenge and the benchmarks are available at \href{https://lidchallenge.github.io}{https://lidchallenge.github.io}.

\end{abstract}

\section{Introduction}
Weakly supervised learning refers to various studies that attempt to address the challenging image recognition tasks by learning from weak or imperfect supervision. Supervised learning methods, including Deep Convolutional Neural Networks (DCNNs), have significantly improved the performance in many problems in the field of computer vision, thanks to the rise of large-scale annotated data set and the advance in computing hardware. However, these supervised learning approaches are notorious ``data-hungry", which makes them are sometimes not practical in many real-world industrial applications. We are often facing the problem that we are not able to acquire enough amount of perfect annotations (\eg, object bounding boxes, and pixel-wise masks) for reliable training models. To address this problem, many efforts in so-called weakly supervised learning approaches have been made to improve the DCNNs training to deviate from traditional paths of supervised learning using imperfect data. For instance, various approaches have proposed new loss functions or novel training schemes. Weakly supervised learning is a popular research direction in Computer Vision and Machine Learning communities. Many research works have been devoted to related topics, leading to the rapid growth of related publications in the top-tier conferences and journals such as CVPR, ICCV, ECCV, NeurIPS, TIP, IJCV, and TPAMI.   

This year, we provide additional annotations for existing benchmarks to enable the weakly-supervised training or evaluation and introduce three challenge tracks to advance the research of weakly-supervised semantic segmentation~\cite{2015-papandreou-weakly,pinheiro2015weakly,xu2015learning,pathak2015constrained,russakovsky2016s,wei2016stc,kolesnikov2016seed,saleh2016built,qi2016augmented,shimoda2016distinct,lin2016scribblesup,wei2016learning,chaudhry2017discovering,khoreva2017simple,wei2017object,kwak2017weakly,xiao2018transferable,tang2018regularized,zhang2019decoupled,zhou2018weakly,ahn2018learning,huang2018weakly,li2018tell,tang2018normalized,wei2018revisiting,hou2018self,Shimoda_2019_ICCV,Zeng_2019_ICCV,Shen_2019_CVPR,Song_2019_CVPR,li2019attention,lee2019ficklenet,wang2019boundary,jiang2019integral,Fan_2020_CVPR,Wang_2020_CVPR,sunmining,luosemi,fanemploying,zhang2020splitting,rahimipairwise,chenweakly,kulharia12356box2seg,arun2020weakly,Olga2020reg}, weakly supervised scene parsing using point supervision~\cite{qian2019weakly} and weakly supervised object localization~\cite{wei2015hcp,oquab2015object,liang2015towards,zhou2016cnnlocalization,bency2016weakly,zhang2016top,cinbis2016weakly,li2016weakly,kantorov2016contextlocnet,jie2017deep,zhu2017soft,kim2017two,singh2017hide,selvaraju2017grad,zhang2018adversarial,zhang2018self,chattopadhay2018grad,Xue_2019_ICCV,Choe_2019_CVPR,choe2020evaluating,zhang2020rethinking,Mai_2020_CVPR,Zhang_2020_CVPR,baerethinking,zhang2020inter,lu2020geometry}, respectively. More details are given in the next section. 

We organize this workshop to investigate principle ways of building industry level AI systems relying on learning from imperfect data. We hope this workshop will attract attention and discussions from both industry and academic people and ease the future research of weakly supervised learning for computer vision.
\section{The LID Challenge}

\subsection{Datasets}

\noindent\textbf{ILSVRC-LID-200} is used in Track 1, aiming to perform object semantic segmentation using image-level annotations as supervision.  This dataset is built upon the object detection track of ImageNet Large Scale Visual Recognition Competition (ILSVRC)~\cite{2009-imagenet}, which totally includes 456,567 training images from 200 categories. By parsing the provided XML files given by~\cite{2009-imagenet}, 349,310 images with object(s) from the 200 categories are left. To facilitate the pixel-level evaluation, we provide pixel-level annotations for 15K images, including 5,000 and 10,000 for validation and testing, respectively. Following previous practices, the mean Intersection-Over-Union (IoU) score over 200 categories is employed as the key evaluation metric.

\vspace{2mm}
\noindent\textbf{ADE20K-LID} is used in Track 2, aiming to learn to perform scene parsing using points-based annotation as supervision. This dataset is built upon the ADE20K dataset~\cite{zhou2017scene}. There are 20,210 images in the training set, 2,000 images in the validation set, and 3,000 images in the testing set. We provide the additional point-based annotations on the training set. In particular, this point-based weakly-supervised setting is firstly provided by~\cite{qian2019weakly}. Following~\cite{qian2019weakly}, we consider 150 meaningful categories and generate the pixel annotation for each independent instance (or region) in each training image. The performances are evaluated by pixel-wise accuracy and mean IoU, which is consistent with the evaluation metrics of the standard ADE20K~\cite{zhou2017scene}.

\vspace{2mm}
\noindent\textbf{ILSVRC} is used in Track 3, aiming to make the classification networks be equipped with the ability of object localization. This dataset is built upon the image classification/localization track of ImageNet Large Scale Visual Recognition Competition (ILSVRC), which includes 1.2 million training images from 1000 categories. Different from previous works to evaluate the performance in an \emph{indirect} way, \ie bounding box, we annotate pixel-level masks of 44,271 images (validation/testing: 23,151/21,120) to facilitate the evaluation to be performed in a \emph{direct} way. These annotations are provided by~\cite{zhang2020rethinking}, where a new evaluation metric, \ie, IoU-Threshold curve, is also introduced. Particularly, the IoU-Threshold curve is obtained by calculating IoU scores with the masks binarized with a wide range of thresholds from 0 to 255. The best IoU score, \ie, Peak-IoU, is used as the key evaluation metric for the comparison. Please refer to \cite{zhang2020rethinking} for more details of the annotation masks and the new evaluation metric. 

\subsection{Rules and Descriptions}

\noindent\textbf{Rules} This year, we issue two strict rules for all the teams
\begin{itemize}
    \item For training, only the images provided in the training set are permitted. Competitors can use the classification models pre-trained on the training set of ILSVRC CLS-LOC to initialize the parameters. However, they CANNOT leverage any datasets with pixel-level annotations. In particular, for Track 1 and Track 3, only the image-level annotations of training images can be leveraged for supervision, and the bounding-box annotations are NOT permitted.
    \item We encourage competitors to design elegant and effective models competing for all the tracks rather than ensembling multiple models. Therefore, we restrict the parameter size of the inference model(s) should be LESS than 150M (slightly more than two DeepLab V3+~\cite{chen2018encoder} models using ResNet-101 as the backbone). The competitors ranked in the Top 3 are required to submit the inference code for verification.
\end{itemize}

\vspace{2mm}
\noindent\textbf{Timeline} The challenge started on Mar 22, 2020, and ended on June 8, 2020. Each participant was allowed a maximum of 5 submissions for the testing split of each track. 

\vspace{2mm}
\noindent\textbf{Participating Teams} We received submissions from 15 teams in total. In particular, the teams from the Computer Vision Lab at ETH Zurich, Tsinghua University, the Vision \& Learning Lab at Seoul National University achieve the 1st place in Track 1, Track 2, and Track 3, respectively.

\subsection{Results and Methods}
\begin{table}[t]

\centering
\caption{Results and rankings of methods submitted to Track 1.}
\scriptsize
\setlength{\tabcolsep}{0.9pt}
\renewcommand{\arraystretch}{0.8}
\resizebox{\linewidth}{!}{
\begin{tabular}{lcccc}
\toprule
Team & Username & Mean IoU & Mean Accuracy & Pixel Accuracy  \\
\midrule
 ETH Zurich & cvl & 45.18 &	59.62 &	80.46 \\
 Seoul National University & VL-task1 & 37.73 & 60.15 &	82.98 \\
 Ukrainian Catholic University & UCU \& SoftServe & 37.34 & 54.87 & 83.64 \\
 - & IOnlyHaveSevenDays	& 36.24	& 68.27	& 84.10 \\
 - & play-njupt	& 31.90	& 46.07	& 82.63 \\
 - & xingxiao & 29.48 & 48.66 & 80.82\\
 - & hagenbreaker & 22.50 & 39.92 & 77.38\\
 - & go-g0 & 19.80 & 38.30 & 76.21\\
 - & lasthours-try & 12.56 & 24.65 & 64.35\\
 - & WH-ljs & 7.79 & 16.59 & 62.52\\
\bottomrule
\end{tabular}
}
\label{tab:track1}
\end{table}
\begin{table}[t]

\centering
\caption{Results and rankings of methods submitted to Track 2.}
\scriptsize
\setlength{\tabcolsep}{0.9pt}
\renewcommand{\arraystretch}{0.8}
\resizebox{\linewidth}{!}{
\begin{tabular}{lcccc}
\toprule
Team & Username & Mean IoU & Mean Accuracy & Pixel Accuracy  \\
\midrule
Tsinghua\&Intel & fromandto & 25.33 & 36.88 & 64.95 \\
\bottomrule
\end{tabular}
}
\label{tab:track2}
\end{table}

\begin{table}[t]
\centering
\caption{Results and rankings of methods submitted to Track 3.}
\scriptsize
\setlength{\tabcolsep}{0.9pt}
\renewcommand{\arraystretch}{0.8}
\resizebox{\linewidth}{!}{
\begin{tabular}{lccc}
\toprule
Team & Username & Peak-IoU & Peak-Threshold  \\
\midrule
Seoul National University & VL-task3 & 63.08 & 24 \\
Mepro-MIC of Beijing Jiaotong U & BJTU-Mepro-MIC  & 61.98 & 35 \\
NUST & LEAP Group of PCA Lab & 61.48 & 7 \\
 - & chohk (wsol-aug) & 52.89 & 36\\
Beijing Normal University & TEN & 48.17	& 42 \\
\bottomrule
\end{tabular}
}
\label{tab:track3}
\end{table}
\begin{figure}[t]
\centering
\includegraphics[width=0.8\linewidth]{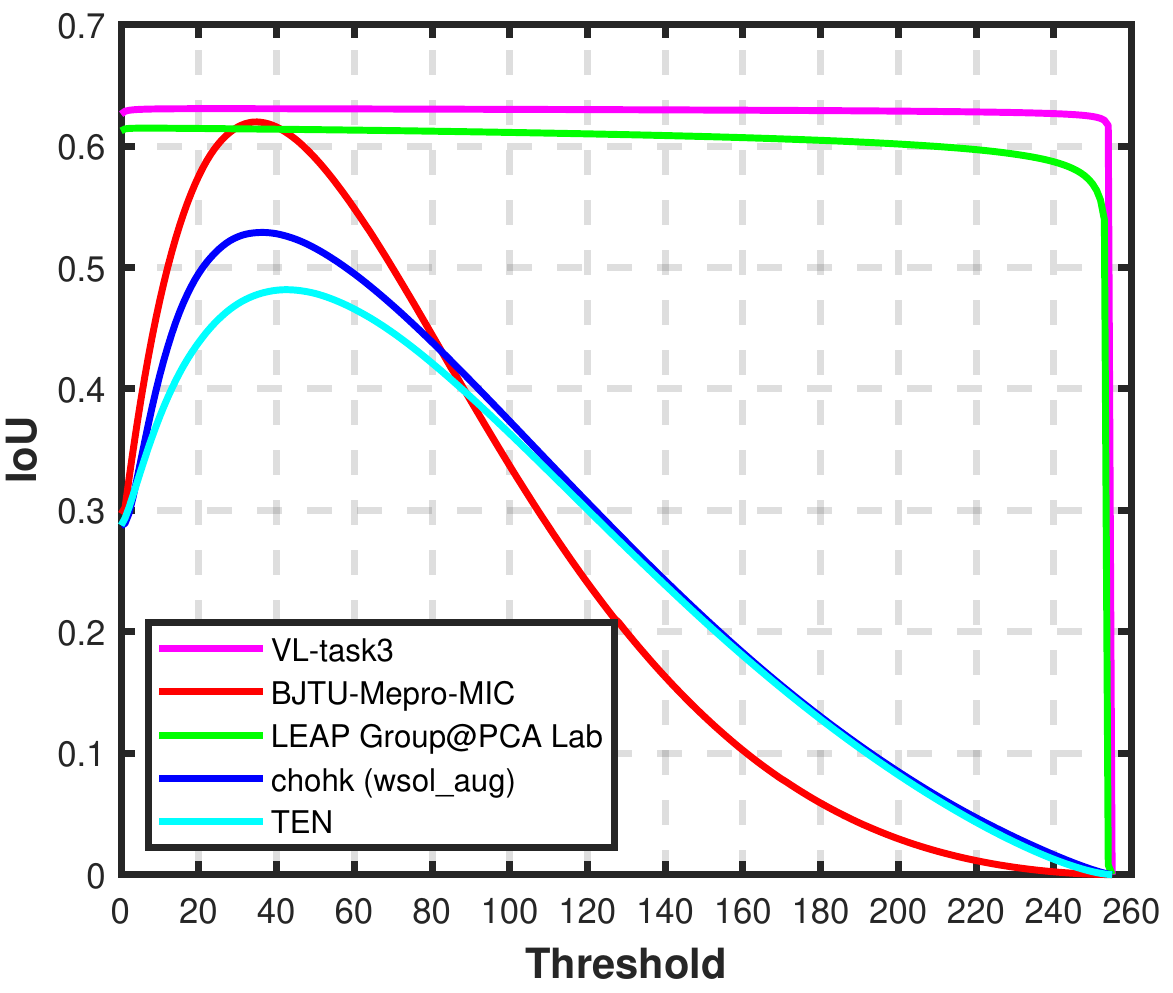}
\caption{The comparison of IoU-Threshold curves of different teams for Track 3.}
\label{fig:curve}   
\end{figure} 

Tables \ref{tab:track1}, \ref{tab:track2}, \ref{tab:track3} show the leaderboard results of Track 1, Track 2 and Track 3, respectively. Particularly, the team from ETH Zurich significantly outperforms others by a large margin in the Track 1. In the Track 3, the top 3 teams achieve similar Peak-IoU scores from 61.48 to 63.08. Furthermore, we demonstrate the comparison of IoU-Threshold curves of 5 teams for Track 3 in Figure~\ref{fig:curve}.

\subsubsection{CVL at ETH Zurich Team \\(Won the 1st place of Track 1)}
\begin{figure}[t]
\centering
\includegraphics[width=\linewidth]{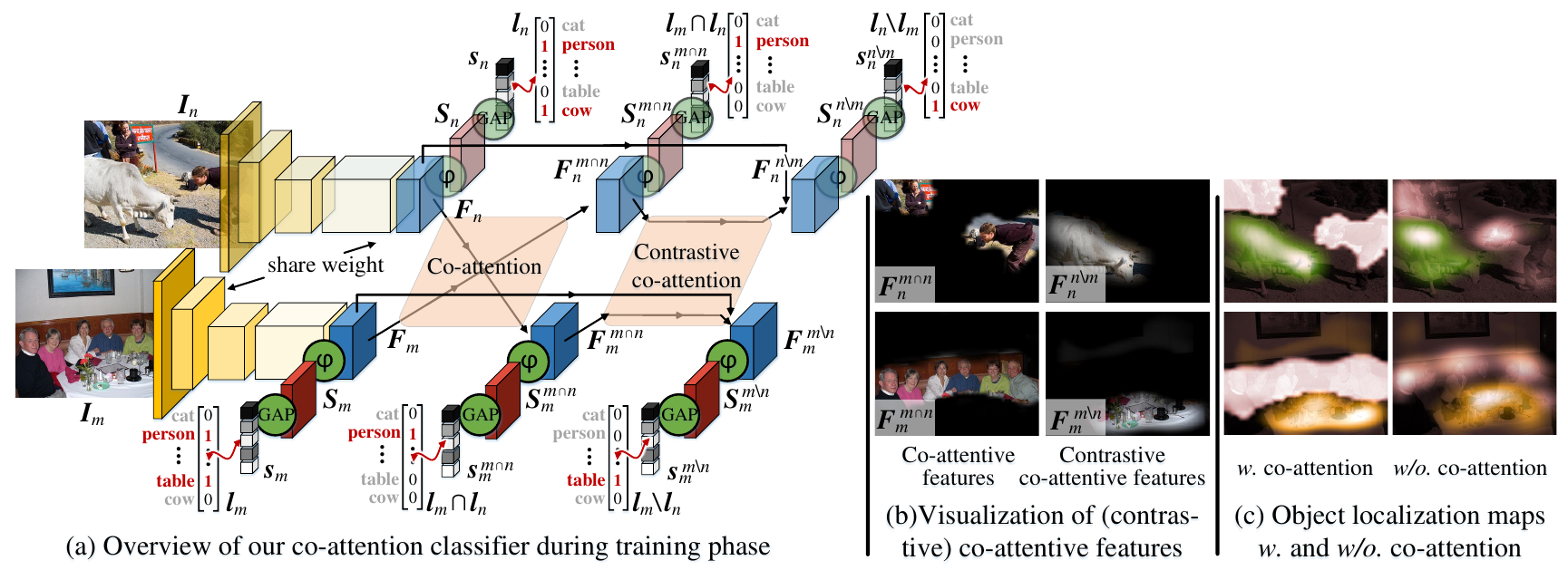}
\caption{An overview of the proposed approach by the team of CVL at ETH Zurich.}
\label{fig:track1-1}   
\end{figure}

The proposed approach adopts cross-image semantic relations for comprehensive object pattern mining. Two neural co-attentions are incorporated into the classifier to complement capture cross-image semantic similarities and differences. In particular, the classifier is equipped with a differentiable co-attention mechanism that addresses semantic homogeneity and difference understanding across training image pairs. More specifically, two kinds of co-attentions are learned in the classifier. The former one aims to capture cross-image common semantics, which enables the classifier to better ground the common semantic labels over the co-attentive regions. The latter, called contrastive co-attention, focuses on the rest, unshared semantics, which helps the classifier better separate semantic patterns of different objects. These two co-attentions work in a cooperative and complimentary
manner, together making the classifier understand object patterns more comprehensively. Another advantage is that the co-attention based classifier learning paradigm brings an efficient data augmentation strategy due to the use of training image pairs. An overview of the proposed approach is shown in Figure~\ref{fig:track1-1}.

\subsubsection{The Machine Learning Lab at Ukrainian Catholic University Team \\(Won the 3rd place of Track 1)}

\begin{figure}[t]
\centering
\includegraphics[width=\linewidth]{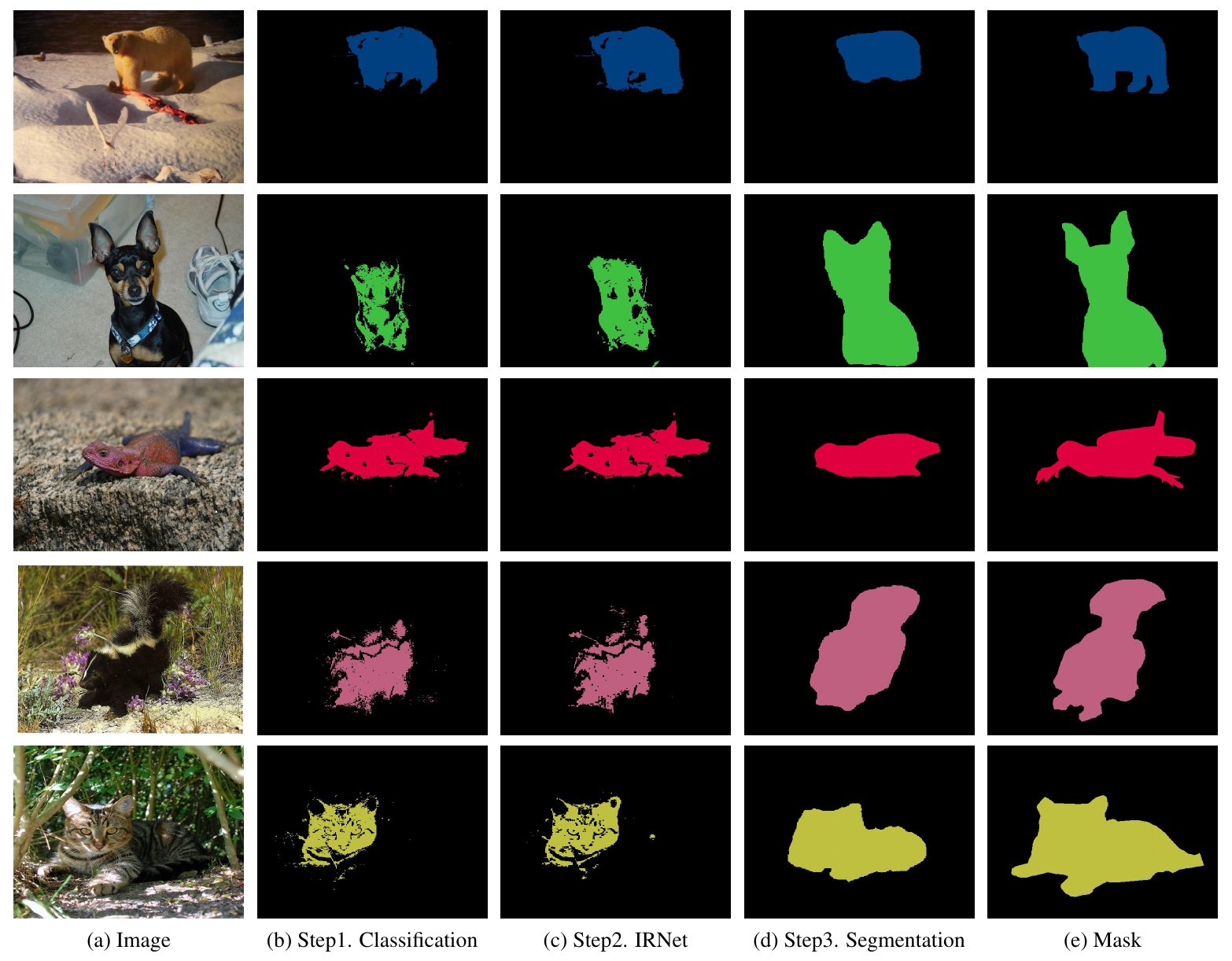}
\caption{The produced localization maps of each consecutive step by the team of Machine Learning Lab at Ukrainian Catholic University.}
\label{fig:track1-3}   
\end{figure}

The approach proposed by this team consists of three consecutive steps. The first two steps extract high-quality pseudo masks from image-level annotated data, which are then used to train a segmentation model on the third step. The presented approach also addresses two problems in the data: class imbalance and missing labels. All these three steps make the proposed approach be capable of segmenting various classes and complex objects using only image-level annotations as supervision. The results produced from different steps are shown in Figure~\ref{fig:track1-3}

\subsubsection{Tsinghua\&Intel Team \\(Won the 1st place of Track 2)}
\begin{figure}[t]
\centering
\includegraphics[width=\linewidth]{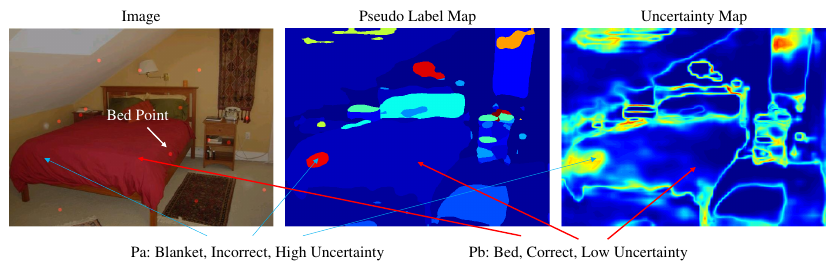}
\caption{Some observations discovered by the team of Machine Learning Lab at Ukrainian Catholic University. Pa: a wrong pseudo label with high uncertainty.
Pb: a correct pseudo label with low uncertainty.}
\label{fig:track2-1}   
\end{figure}

The team reveals two critical issues existing in the current state-of-the-art method~\cite{qian2019weakly}: 1) it relies upon softmax outputs, or say logits. It is known that logits can be over-confident upon the wrong prediction; 2) harvesting pseudo labels using logits would introduce thresholds, and it is very time-consuming to tune thresholds for modern deep networks. Some observations are shown in Figure~\ref{fig:track2-1}.

To tackle these issues, the proposed approach builds upon uncertainty measures instead of logits and is free of threshold tuning, which is motivated by a large-scale analysis of the distribution of uncertainty measures using strong models and challenging databases. This analysis leads to the discovery
of a statistical phenomenon called uncertainty mixture. Inspired by this discovery, this team proposes to decompose the distribution of uncertainty measures with a Gamma mixture model, leading to a principled method to harvest reliable pseudo labels. Beyond that, the team assumes the uncertainty
measures for labeled points are always drawn from a particular component. This amounts to a regularized Gamma mixture model. They provide a thorough theoretical analysis of this model, showing that it can be solved with an EM-style algorithm with a convergence guarantee.

\subsubsection{Vision\&Learning Lab at Seoul National University Team \\(Won the 1st and 2nd places for Track 3 and Track 1)}
\begin{figure}[t]
\centering
\includegraphics[width=\linewidth]{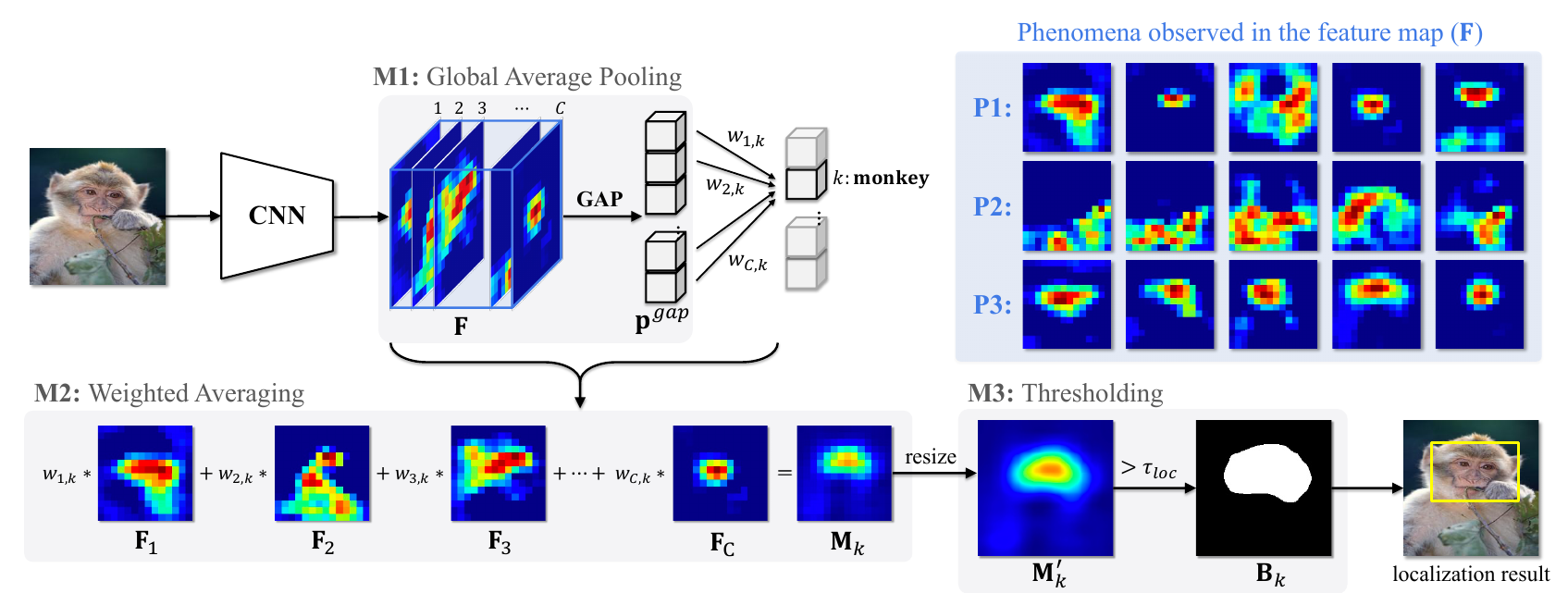}
\caption{An overview of the proposed approach by the team of Vision\&Learning Lab at Seoul National University.}
\label{fig:track3-1}   
\end{figure}

This team demonstrates the popular class activation maps~\cite{zhou2016cnnlocalization} suffers from three fundamental issues: (i) the bias of GAP to assign a higher weight to a channel with a small activation area, (ii) negatively weighted activations inside the object regions, and (iii) instability from the use of the maximum value of a class activation map as a thresholding reference. They collectively cause the problem that the localization prediction to be highly limited to the small region of an object. The proposed approach incorporates three simple but robust techniques that alleviate the regarding problems, including thresholded average pooling, negative weight clamping, and percentile as a thresholding
standard. More details can be found in Figure~\ref{fig:track3-1}.

\subsubsection{Mepro at Beijing Jiaotong University Team \\(Won the 2nd place of Track 3)}
\begin{figure}[t]
\centering
\includegraphics[width=\linewidth]{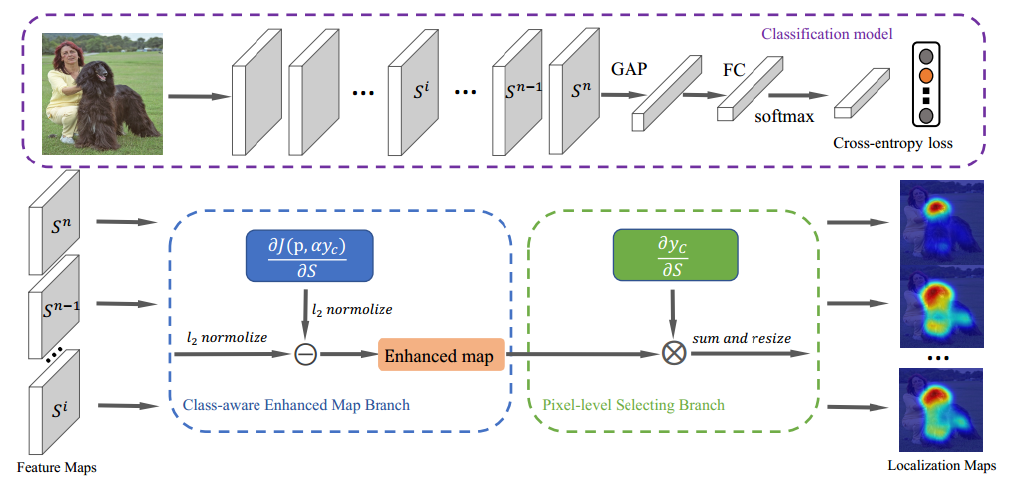}
\caption{An overview of the Dual-Gradients Localization (DGL) framework proposed by the team of Mepro at Beijing Jiaotong University.}
\label{fig:track3-2}   
\end{figure}

The proposed method achieves localization on any convolutional layer of a classification model by exploiting two kinds of gradients, called the Dual-Gradients Localization (DGL) framework. DGL framework is developed based on two branches: 1) Pixel-level Class Selection, leveraging gradients of target class to identify the correlation ratio of pixels to the target class within any convolutional feature maps, and 2) Class-aware Enhanced Map, utilizing gradients of the classification loss function to mine entire target object regions, which would not damage classification performance. The proposed architecture is shown in Figure~\ref{fig:track3-2}.

To acquire the integral object regions, gradients of classification loss function are used to enhance the information of the specific class on any convolutional layer. DGL framework demonstrates that any convolutional layer of the classification model has the localization ability, and some layers have better performance than the last convolutional layer in an offline manner.

\subsubsection{PCA Lab at Nanjing University of Science and Technology Team \\(Won the 3rd place of Track 3)}

\begin{figure*}[t]
\centering
\includegraphics[width=0.7\linewidth]{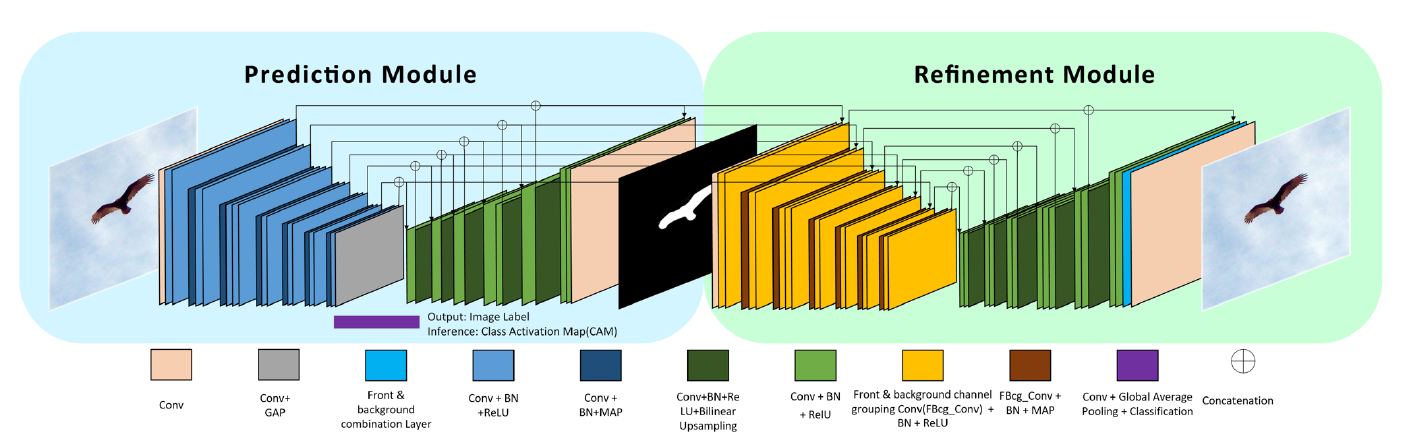}
\caption{An overview of the framework proposed by the team of PCA Lab at Nanjing University of Science and Technology.}
\label{fig:track3-3}   
\end{figure*} 

The proposed model is composed of two auto-encoders, as shown in Figure~\ref{fig:track3-3}. In the first auto-encoder, a classifier is trained with the global average pooling by using image-level annotations as supervision. The learned classifier is further applied to obtain Class Activation Maps (CAMs) according to~\cite{zhou2016cnnlocalization}. Then, the team uses the binary images generated by CAMs as pseudo-pixel-level annotations to conduct binary classification. After the first decoder, a bilinear upsampling operation is further applied to get the binary image with the same size as the raw image.

In the second auto-encoder, the team aims to recover the binary image to the raw image to obtain the refined binary output image. The output binary image of the first decoder is used as the input of the second encoder. The binary image is compounded with a raw image and is then divided into front and background. The front part and the background part are operated in different channels of a layer independently and then are fused to the final prediction image. To be specific, they transfer front and background images in all layers into binary images with a threshold, add front and background binary images together, and then fuse the binary images generated by all layers. As a result, we may get the refined binary output images, which are used as pseudo annotations for the next iteration. The above process iterates until convergence.

\subsubsection{Beijing Normal University \\(Won the 5th place of Track 3)}
This team simply apply the approach proposed in~\cite{Choe_2019_CVPR} to compete the Track 3.

\paragraph{Acknowledgments}
We thank our sponsor of the \href{https://github.com/PaddlePaddle/Paddle}{PaddlePaddle} Team from Baidu.

{\small
\bibliographystyle{ieee_fullname}
\bibliography{egbib}

\begin{thebibliography}{10}\itemsep=-1pt

\bibitem{ahn2018learning}
Jiwoon Ahn and Suha Kwak.
\newblock Learning pixel-level semantic affinity with image-level supervision
  for weakly supervised semantic segmentation.
\newblock In {\em {CVPR}}, pages 4981--4990, 2018.

\bibitem{arun2020weakly}
Aditya Arun, CV Jawahar, and M~Pawan Kumar.
\newblock Weakly supervised instance segmentation by learning annotation
  consistent instances.
\newblock In {\em {ECCV}}, 2020.

\bibitem{baerethinking}
Wonho Bae, Junhyug Noh, and Gunhee Kim.
\newblock Rethinking class activation mapping for weakly supervised object
  localization.
\newblock In {\em {ECCV}}, 2020.

\bibitem{bency2016weakly}
Archith~John Bency, Heesung Kwon, Hyungtae Lee, S Karthikeyan, and BS
  Manjunath.
\newblock Weakly supervised localization using deep feature maps.
\newblock In {\em eccv}, pages 714--731, 2016.

\bibitem{chattopadhay2018grad}
Aditya Chattopadhay, Anirban Sarkar, Prantik Howlader, and Vineeth~N
  Balasubramanian.
\newblock Grad-cam++: Generalized gradient-based visual explanations for deep
  convolutional networks.
\newblock In {\em WACV}, pages 839--847, 2018.

\bibitem{chaudhry2017discovering}
Arslan Chaudhry, Puneet~K Dokania, and Philip~HS Torr.
\newblock Discovering class-specific pixels for weakly-supervised semantic
  segmentation.
\newblock {\em arXiv preprint arXiv:1707.05821}, 2017.

\bibitem{chenweakly}
Liyi Chen, Weiwei Wu, Chenchen Fu, Xiao Han, and Yuntao Zhang.
\newblock Weakly supervised semantic segmentation with boundary exploration.
\newblock In {\em {ECCV}}, 2020.

\bibitem{chen2018encoder}
Liang-Chieh Chen, Yukun Zhu, George Papandreou, Florian Schroff, and Hartwig
  Adam.
\newblock Encoder-decoder with atrous separable convolution for semantic image
  segmentation.
\newblock In {\em {ECCV}}, pages 801--818, 2018.

\bibitem{choe2020evaluating}
Junsuk Choe, Seong~Joon Oh, Seungho Lee, Sanghyuk Chun, Zeynep Akata, and
  Hyunjung Shim.
\newblock Evaluating weakly supervised object localization methods right.
\newblock In {\em cvpr}, 2020.

\bibitem{Choe_2019_CVPR}
Junsuk Choe and Hyunjung Shim.
\newblock Attention-based dropout layer for weakly supervised object
  localization.
\newblock In {\em {CVPR}}, June 2019.

\bibitem{cinbis2016weakly}
Ramazan~Gokberk Cinbis, Jakob Verbeek, and Cordelia Schmid.
\newblock Weakly supervised object localization with multi-fold multiple
  instance learning.
\newblock {\em {TPAMI}}, 39(1):189--203, 2016.

\bibitem{2009-imagenet}
Jia Deng, Wei Dong, Richard Socher, Li-Jia Li, Kai Li, and Li Fei-Fei.
\newblock Imagenet: A large-scale hierarchical image database.
\newblock In {\em {CVPR}}, pages 248--255, 2009.

\bibitem{Fan_2020_CVPR}
Junsong Fan, Zhaoxiang Zhang, Chunfeng Song, and Tieniu Tan.
\newblock Learning integral objects with intra-class discriminator for
  weakly-supervised semantic segmentation.
\newblock In {\em {CVPR}}, 2020.

\bibitem{fanemploying}
Junsong Fan, Zhaoxiang Zhang, and Tieniu Tan.
\newblock Employing multi-estimations for weakly-supervised semantic
  segmentation.
\newblock In {\em {ECCV}}, 2020.

\bibitem{hou2018self}
Qibin Hou, PengTao Jiang, Yunchao Wei, and Ming-Ming Cheng.
\newblock Self-erasing network for integral object attention.
\newblock In {\em {NIPS}}, pages 549--559, 2018.

\bibitem{huang2018weakly}
Zilong Huang, Xinggang Wang, Jiasi Wang, Wenyu Liu, and Jingdong Wang.
\newblock Weakly-supervised semantic segmentation network with deep seeded
  region growing.
\newblock In {\em {CVPR}}, pages 7014--7023, 2018.

\bibitem{jiang2019integral}
Peng-Tao Jiang, Qibin Hou, Yang Cao, Ming-Ming Cheng, Yunchao Wei, and Hong-Kai
  Xiong.
\newblock Integral object mining via online attention accumulation.
\newblock In {\em {ICCV}}, pages 2070--2079, 2019.

\bibitem{jie2017deep}
Zequn Jie, Yunchao Wei, Xiaojie Jin, Jiashi Feng, and Wei Liu.
\newblock Deep self-taught learning for weakly supervised object localization.
\newblock In {\em {CVPR}}, 2017.

\bibitem{kantorov2016contextlocnet}
Vadim Kantorov, Maxime Oquab, Minsu Cho, and Ivan Laptev.
\newblock Contextlocnet: Context-aware deep network models for weakly
  supervised localization.
\newblock In {\em {ECCV}}, pages 350--365, 2016.

\bibitem{khoreva2017simple}
Anna Khoreva, Rodrigo Benenson, Jan~Hendrik Hosang, Matthias Hein, and Bernt
  Schiele.
\newblock Simple does it: Weakly supervised instance and semantic segmentation.
\newblock In {\em {CVPR}}, volume~1, page~3, 2017.

\bibitem{kim2017two}
Dahun Kim, Donghyeon Cho, Donggeun Yoo, and In So~Kweon.
\newblock Two-phase learning for weakly supervised object localization.
\newblock In {\em {ICCV}}, pages 3534--3543, 2017.

\bibitem{kolesnikov2016seed}
Alexander Kolesnikov and Christoph~H Lampert.
\newblock Seed, expand and constrain: Three principles for weakly-supervised
  image segmentation.
\newblock In {\em {ECCV}}, pages 695--711, 2016.

\bibitem{kulharia12356box2seg}
Viveka Kulharia, Siddhartha Chandra, Amit Agrawal, Philip Torr, and Ambrish
  Tyagi.
\newblock Box2seg: Attention weighted loss and discriminative feature learning
  for weakly supervised segmentation.
\newblock In {\em {ECCV}}, 2020.

\bibitem{kwak2017weakly}
Suha Kwak, Seunghoon Hong, Bohyung Han, et~al.
\newblock Weakly supervised semantic segmentation using superpixel pooling
  network.
\newblock In {\em AAAI}, pages 4111--4117, 2017.

\bibitem{lee2019ficklenet}
Jungbeom Lee, Eunji Kim, Sungmin Lee, Jangho Lee, and Sungroh Yoon.
\newblock Ficklenet: Weakly and semi-supervised semantic image segmentation
  using stochastic inference.
\newblock In {\em {CVPR}}, pages 5267--5276, 2019.

\bibitem{li2016weakly}
Dong Li, Jia-Bin Huang, Yali Li, Shengjin Wang, and Ming-Hsuan Yang.
\newblock Weakly supervised object localization with progressive domain
  adaptation.
\newblock In {\em {CVPR}}, pages 3512--3520, 2016.

\bibitem{li2018tell}
Kunpeng Li, Ziyan Wu, Kuan-Chuan Peng, Jan Ernst, and Yun Fu.
\newblock Tell me where to look: Guided attention inference network.
\newblock In {\em {CVPR}}, pages 9215--9223, 2018.

\bibitem{li2019attention}
Kunpeng Li, Yulun Zhang, Kai Li, Yuanyuan Li, and Yun Fu.
\newblock Attention bridging network for knowledge transfer.
\newblock In {\em {CVPR}}, pages 5198--5207, 2019.

\bibitem{liang2015towards}
Xiaodan Liang, Si Liu, Yunchao Wei, Luoqi Liu, Liang Lin, and Shuicheng Yan.
\newblock Towards computational baby learning: A weakly-supervised approach for
  object detection.
\newblock In {\em {ICCV}}, pages 999--1007, 2015.

\bibitem{lin2016scribblesup}
Di Lin, Jifeng Dai, Jiaya Jia, Kaiming He, and Jian Sun.
\newblock Scribblesup: Scribble-supervised convolutional networks for semantic
  segmentation.
\newblock {\em {CVPR}}, 2016.

\bibitem{lu2020geometry}
Weizeng Lu, Xi Jia, Weicheng Xie, Linlin Shen, Yicong Zhou, and Jinming Duan.
\newblock Geometry constrained weakly supervised object localization.
\newblock In {\em {ECCV}}, 2020.

\bibitem{luosemi}
Wenfeng Luo and Meng Yang.
\newblock Semi-supervised semantic segmentation via strong-weak dual-branch
  network.
\newblock In {\em {ECCV}}, 2020.

\bibitem{Mai_2020_CVPR}
Jinjie Mai, Meng Yang, and Wenfeng Luo.
\newblock Erasing integrated learning: A simple yet effective approach for
  weakly supervised object localization.
\newblock In {\em {CVPR}}, 2020.

\bibitem{oquab2015object}
Maxime Oquab, L{\'e}on Bottou, Ivan Laptev, and Josef Sivic.
\newblock Is object localization for free?-weakly-supervised learning with
  convolutional neural networks.
\newblock In {\em {CVPR}}, pages 685--694, 2015.

\bibitem{2015-papandreou-weakly}
George Papandreou, Liang-Chieh Chen, Kevin Murphy, and Alan~L Yuille.
\newblock Weakly-and semi-supervised learning of a dcnn for semantic image
  segmentation.
\newblock In {\em {ICCV}}, 2015.

\bibitem{pathak2015constrained}
Deepak Pathak, Philipp Krahenbuhl, and Trevor Darrell.
\newblock Constrained convolutional neural networks for weakly supervised
  segmentation.
\newblock In {\em {ICCV}}, pages 1796--1804, 2015.

\bibitem{pinheiro2015weakly}
Pedro~O Pinheiro and Ronan Collobert.
\newblock From image-level to pixel-level labeling with convolutional networks.
\newblock In {\em {CVPR}}, 2015.

\bibitem{qi2016augmented}
Xiaojuan Qi, Zhengzhe Liu, Jianping Shi, Hengshuang Zhao, and Jiaya Jia.
\newblock Augmented feedback in semantic segmentation under image level
  supervision.
\newblock In {\em {ECCV}}, pages 90--105, 2016.

\bibitem{qian2019weakly}
Rui Qian, Yunchao Wei, Honghui Shi, Jiachen Li, Jiaying Liu, and Thomas Huang.
\newblock Weakly supervised scene parsing with point-based distance metric
  learning.
\newblock In {\em AAAI}, volume~33, pages 8843--8850, 2019.

\bibitem{rahimipairwise}
Amir Rahimi, Amirreza Shaban, Thalaiyasingam Ajanthan, Richard Hartley, and
  Byron Boots.
\newblock Pairwise similarity knowledge transfer for weakly supervised object
  localization.
\newblock In {\em {ECCV}}, 2020.

\bibitem{russakovsky2016s}
O Russakovsky, AL Bearman, V Ferrari, and L Fei-Fei.
\newblock What’s the point: Semantic segmentation with point supervision.
\newblock In {\em {ECCV}}, pages 549--565, 2016.

\bibitem{saleh2016built}
Fatemehsadat Saleh, Mohammad Sadegh~Ali Akbarian, Mathieu Salzmann, Lars
  Petersson, Stephen Gould, and Jose~M Alvarez.
\newblock Built-in foreground/background prior for weakly-supervised semantic
  segmentation.
\newblock In {\em {ECCV}}, pages 413--432, 2016.

\bibitem{selvaraju2017grad}
Ramprasaath~R Selvaraju, Michael Cogswell, Abhishek Das, Ramakrishna Vedantam,
  Devi Parikh, and Dhruv Batra.
\newblock Grad-cam: Visual explanations from deep networks via gradient-based
  localization.
\newblock In {\em {ICCV}}, pages 618--626, 2017.

\bibitem{Shen_2019_CVPR}
Yunhang Shen, Rongrong Ji, Yan Wang, Yongjian Wu, and Liujuan Cao.
\newblock Cyclic guidance for weakly supervised joint detection and
  segmentation.
\newblock In {\em {CVPR}}, 2019.

\bibitem{shimoda2016distinct}
Wataru Shimoda and Keiji Yanai.
\newblock Distinct class-specific saliency maps for weakly supervised semantic
  segmentation.
\newblock In {\em {ECCV}}, pages 218--234, 2016.

\bibitem{Shimoda_2019_ICCV}
Wataru Shimoda and Keiji Yanai.
\newblock Self-supervised difference detection for weakly-supervised semantic
  segmentation.
\newblock In {\em {ICCV}}, 2019.

\bibitem{singh2017hide}
Krishna~Kumar Singh and Yong~Jae Lee.
\newblock Hide-and-seek: Forcing a network to be meticulous for
  weakly-supervised object and action localization.
\newblock In {\em {ICCV}}, pages 3544--3553, 2017.

\bibitem{Song_2019_CVPR}
Chunfeng Song, Yan Huang, Wanli Ouyang, and Liang Wang.
\newblock Box-driven class-wise region masking and filling rate guided loss for
  weakly supervised semantic segmentation.
\newblock In {\em {CVPR}}, 2019.

\bibitem{sunmining}
Guolei Sun, Wenguan Wang, Jifeng Dai, and Luc Van~Gool.
\newblock Mining cross-image semantics for weakly supervised semantic
  segmentation.
\newblock In {\em {ECCV}}, 2020.

\bibitem{tang2018normalized}
Meng Tang, Abdelaziz Djelouah, Federico Perazzi, Yuri Boykov, and Christopher
  Schroers.
\newblock Normalized cut loss for weakly-supervised cnn segmentation.
\newblock In {\em {CVPR}}, 2018.

\bibitem{tang2018regularized}
Meng Tang, Federico Perazzi, Abdelaziz Djelouah, Ismail Ben~Ayed, Christopher
  Schroers, and Yuri Boykov.
\newblock On regularized losses for weakly-supervised cnn segmentation.
\newblock In {\em {ECCV}}, pages 507--522, 2018.

\bibitem{Olga2020reg}
Olga Veksler.
\newblock Regularized loss for weakly supervised single class semantic
  segmentation.
\newblock In {\em {ECCV}}, 2020.

\bibitem{wang2019boundary}
Bin Wang, Guojun Qi, Sheng Tang, Tianzhu Zhang, Yunchao Wei, Linghui Li, and
  Yongdong Zhang.
\newblock Boundary perception guidance: a scribble-supervised semantic
  segmentation approach.
\newblock In {\em IJCAI}, pages 3663--3669, 2019.

\bibitem{Wang_2020_CVPR}
Yude Wang, Jie Zhang, Meina Kan, Shiguang Shan, and Xilin Chen.
\newblock Self-supervised equivariant attention mechanism for weakly supervised
  semantic segmentation.
\newblock In {\em {CVPR}}, 2020.

\bibitem{wei2017object}
Yunchao Wei, Jiashi Feng, Xiaodan Liang, Ming-Ming Cheng, Yao Zhao, and
  Shuicheng Yan.
\newblock Object region mining with adversarial erasing: A simple
  classification to semantic segmentation approach.
\newblock In {\em {CVPR}}, 2017.

\bibitem{wei2016learning}
Yunchao Wei, Xiaodan Liang, Yunpeng Chen, Zequn Jie, Yanhui Xiao, Yao Zhao, and
  Shuicheng Yan.
\newblock Learning to segment with image-level annotations.
\newblock {\em Pattern Recognition}, 2016.

\bibitem{wei2016stc}
Yunchao Wei, Xiaodan Liang, Yunpeng Chen, Xiaohui Shen, Ming-Ming Cheng, Jiashi
  Feng, Yao Zhao, and Shuicheng Yan.
\newblock Stc: A simple to complex framework for weakly-supervised semantic
  segmentation.
\newblock {\em {TPAMI}}, 2016.

\bibitem{wei2015hcp}
Yunchao Wei, Wei Xia, Min Lin, Junshi Huang, Bingbing Ni, Jian Dong, Yao Zhao,
  and Shuicheng Yan.
\newblock Hcp: A flexible cnn framework for multi-label image classification.
\newblock {\em {TPAMI}}, (9):1901--1907, 2015.

\bibitem{wei2018revisiting}
Yunchao Wei, Huaxin Xiao, Honghui Shi, Zequn Jie, Jiashi Feng, and Thomas~S
  Huang.
\newblock Revisiting dilated convolution: A simple approach for weakly-and
  semi-supervised semantic segmentation.
\newblock In {\em {CVPR}}, pages 7268--7277, 2018.

\bibitem{xiao2018transferable}
Huaxin Xiao, Yunchao Wei, Yu Liu, Maojun Zhang, and Jiashi Feng.
\newblock Transferable semi-supervised semantic segmentation.
\newblock In {\em AAAI}, 2018.

\bibitem{xu2015learning}
Jia Xu, Alexander~G Schwing, and Raquel Urtasun.
\newblock Learning to segment under various forms of weak supervision.
\newblock In {\em {CVPR}}, 2015.

\bibitem{Xue_2019_ICCV}
Haolan Xue, Chang Liu, Fang Wan, Jianbin Jiao, Xiangyang Ji, and Qixiang Ye.
\newblock Danet: Divergent activation for weakly supervised object
  localization.
\newblock In {\em {ICCV}}, October 2019.

\bibitem{Zeng_2019_ICCV}
Yu Zeng, Yunzhi Zhuge, Huchuan Lu, and Lihe Zhang.
\newblock Joint learning of saliency detection and weakly supervised semantic
  segmentation.
\newblock In {\em {ICCV}}, October 2019.

\bibitem{Zhang_2020_CVPR}
Chen-Lin Zhang, Yun-Hao Cao, and Jianxin Wu.
\newblock Rethinking the route towards weakly supervised object localization.
\newblock In {\em {CVPR}}, 2020.

\bibitem{zhang2016top}
Jianming Zhang, Zhe Lin, Jonathan Brandt, Xiaohui Shen, and Stan Sclaroff.
\newblock Top-down neural attention by excitation backprop.
\newblock In {\em {ECCV}}, pages 543--559, 2016.

\bibitem{zhang2019decoupled}
Tianyi Zhang, Guosheng Lin, Jianfei Cai, Tong Shen, Chunhua Shen, and Alex~C
  Kot.
\newblock Decoupled spatial neural attention for weakly supervised semantic
  segmentation.
\newblock 21(11):2930--2941, 2019.

\bibitem{zhang2020splitting}
Tianyi Zhang, Guosheng Lin, Weide Liu, Jianfei Cai, and Alex Kot.
\newblock Splitting vs. merging: Mining object regions with discrepancy and
  intersection loss for weakly supervised semantic segmentation.
\newblock In {\em {ECCV}}, 2020.

\bibitem{zhang2018adversarial}
Xiaolin Zhang, Yunchao Wei, Jiashi Feng, Yi Yang, and Thomas Huang.
\newblock Adversarial complementary learning for weakly supervised object
  localization.
\newblock In {\em {CVPR}}, 2018.

\bibitem{zhang2018self}
Xiaolin Zhang, Yunchao Wei, Guoliang Kang, Yi Yang, and Thomas Huang.
\newblock Self-produced guidance for weakly-supervised object localization.
\newblock In {\em {ECCV}}, 2018.

\bibitem{zhang2020inter}
Xiaolin Zhang, Yunchao Wei, and Yi Yang.
\newblock Inter-image communication for weakly supervised localization.
\newblock In {\em {ECCV}}, 2020.

\bibitem{zhang2020rethinking}
Xiaolin Zhang, Yunchao Wei, Yi Yang, and Fei Wu.
\newblock Rethinking localization map: Towards accurate object perception with
  self-enhancement maps.
\newblock {\em arXiv preprint arXiv:2006.05220}, 2020.

\bibitem{zhou2016cnnlocalization}
Bolei Zhou, Aditya Khosla, Agata Lapedriza, Aude Oliva, and Antonio Torralba.
\newblock Learning deep features for discriminative localization.
\newblock In {\em {CVPR}}, pages 2921--2929, 2016.

\bibitem{zhou2017scene}
Bolei Zhou, Hang Zhao, Xavier Puig, Sanja Fidler, Adela Barriuso, and Antonio
  Torralba.
\newblock Scene parsing through ade20k dataset.
\newblock In {\em {CVPR}}, pages 633--641, 2017.

\bibitem{zhou2018weakly}
Yanzhao Zhou, Yi Zhu, Qixiang Ye, Qiang Qiu, and Jianbin Jiao.
\newblock Weakly supervised instance segmentation using class peak response.
\newblock In {\em {CVPR}}, pages 3791--3800, 2018.

\bibitem{zhu2017soft}
Yi Zhu, Yanzhao Zhou, Qixiang Ye, Qiang Qiu, and Jianbin Jiao.
\newblock Soft proposal networks for weakly supervised object localization.
\newblock {\em arXiv preprint arXiv:1709.01829}, 2017.

\end{thebibliography}
}

\end{document}